\documentclass[sigconf]{acmart}

\usepackage{booktabs}
\usepackage{multirow}
\usepackage[ruled,linesnumbered]{algorithm2e} 
\AtBeginDocument{%
  }

\setcopyright{acmlicensed}
\copyrightyear{2018}
\acmYear{2018}
\acmDOI{XXXXXXX.XXXXXXX}
\acmConference[Conference acronym 'XX]{Make sure to enter the correct
  conference title from your rights confirmation email}{June 03--05,
  2018}{Woodstock, NY}
\acmISBN{978-1-4503-XXXX-X/2018/06}

\begin{document}

\title{A3R: Agentic Affordance Reasoning via Cross-Dimensional Evidence in 3D Gaussian Scenes
}

\author{Di Li}
\email{dili@stu.xidian.edu.cn}
\affiliation{%
  \institution{Xidian University}
  \country{China}
}

\author{Jie Feng}
\authornote{Corresponding author.}
\email{jiefeng0109@163.com}
\affiliation{%
  \institution{Xidian University}
  \country{China}
}

\author{Guanbin Li}
\email{liguanbin@mail.sysu.edu.cn}
\affiliation{%
  \institution{Sun Yat-sen University}
  \country{China}
}

\author{Ronghua Shang}
\email{rhshang@mail.xidian.edu.cn}
\affiliation{%
  \institution{Xidian University}
  \country{China}
}

\author{Yuhui Zheng}
\email{zhengyh@vip.126.com}
\affiliation{%
  \institution{Qinghai Normal University}
  \country{China}
}

\author{Weisheng Dong}
\email{wsdong@mail.xidian.edu.cn}
\affiliation{%
  \institution{Xidian University}
  \country{China}
}

\author{Guangming Shi}
\email{gmshi@xidian.edu.cn}
\affiliation{%
  \institution{Xidian University}
  \country{China}
}

\renewcommand{\shortauthors}{Trovato et al.}

\begin{abstract}
Affordance reasoning in 3D Gaussian scenes aims to identify the region that supports the action specified by a given text instruction in complex environments.
Existing methods typically cast this problem as one-shot prediction from static scene observations, assuming sufficient evidence is already available for reasoning. However, in complex 3D scenes, many failure cases arise not from weak prediction capacity, but from incomplete task-relevant evidence under fixed observations. To address this limitation, we reformulate fine-grained affordance reasoning as a sequential evidence acquisition process, where ambiguity is progressively reduced through complementary 3D geometric and 2D semantic evidence. Building on this formulation, we propose A3R, an agentic affordance reasoning framework that enables an MLLM-based policy to iteratively select evidence acquisition actions and update the affordance belief through cross-dimensional evidence acquisition. To optimize such sequential decision making, we further introduce a GRPO-based policy learning strategy that improves evidence acquisition efficiency and reasoning accuracy. Extensive experiments on scene-level benchmarks show that A3R consistently surpasses static one-shot baselines, demonstrating the advantage of agentic cross-dimensional evidence acquisition for fine-grained affordance reasoning in complex 3D Gaussian scenes.

\end{abstract}


\begin{teaserfigure}
  \includegraphics[width=\textwidth]{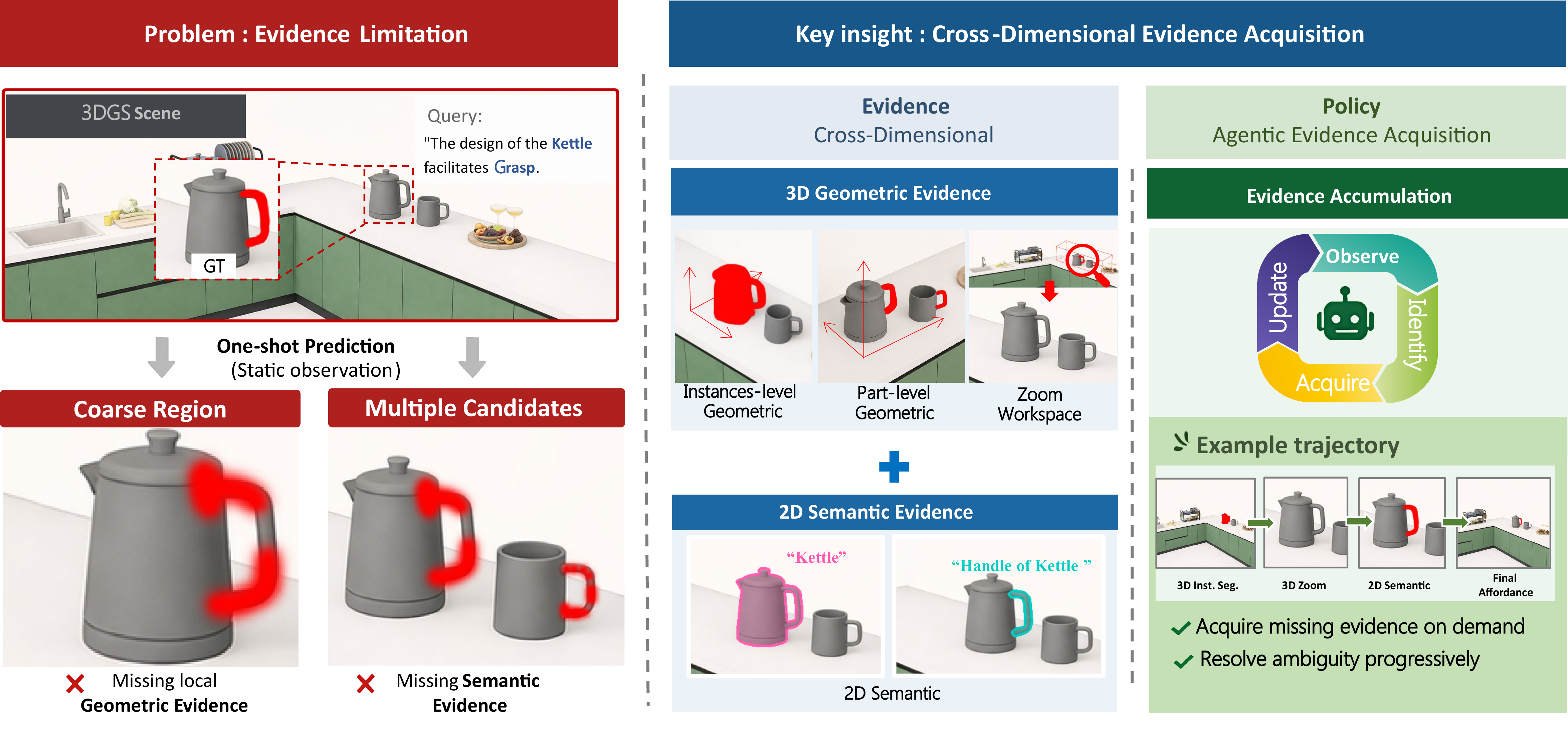}
\caption{
Affordance reasoning in complex 3DGS scenes is often limited by insufficient task-relevant evidence under static observation. Our framework addresses this issue through agentic cross-dimensional evidence acquisition, which progressively gathers complementary 3D geometric and 2D semantic evidence to reduce ambiguity and produce accurate fine-grained predictions.
}
\Description{Enjoying}
    \label{fig:Overview}
  \label{fig:teaser}
\end{teaserfigure}

\maketitle

\section{Introduction}

\begin{table*}[t]
\centering
\small
\setlength{\tabcolsep}{7pt}
\caption{Comparison with representative affordance reasoning methods. Our method differs from prior work by performing agentic cross-dimensional evidence acquisition in 3DGS scenes with both 2D and 3D tools.}
\begin{tabular}{@{}l c c c c@{}}
\toprule
Method & Venue & Scene Rep. & Inference Mode & Cross-Dimensional Evidence \\
\midrule
3DAffordanceNet~\cite{deng20213d}          & CVPR'21    & Point cloud & Static one-shot                  & None \\
IAGNet~\cite{yang2023grounding}            & ICCV'23    & Point cloud & Static one-shot                  & Static Fusion \\
PointRefer~\cite{li2024laso}               & CVPR'24    & Point cloud & Static one-shot                  & None \\
AffordSplatNet~\cite{wei20253daffordsplat} & MM'25      & 3DGS        & Static one-shot                  & Static Fusion \\
SeqAffordSplat~\cite{li2025seqaffordsplat} & arXiv'25   & 3DGS        & Static one-shot                  & Static Fusion \\
AffordBot~\cite{wang2025affordbot}         & NeurIPS'25 & Point cloud & Active viewpoint selection       & View-mediated \\
A4-Agent~\cite{zhang2025a4}                & arXiv'25   & Image       & Agentic 2D tool use              & 2D only \\
\midrule
\textbf{Ours}                              & -          & \textbf{3DGS} & \textbf{Agentic 2D/3D evidence acquisition} & \textbf{Dynamic Querying} \\
\bottomrule
\end{tabular}

\label{tab:rw_comparison}
\end{table*}

Affordance reasoning aims to localize the object or region in a scene that supports a text-specified interaction, serving as a critical bridge between visual perception and physical action in embodied intelligence~\cite{yang2023grounding,li2024laso,zhu2025grounding}. 
Unlike conventional recognition tasks that focus on identifying what is present, affordance reasoning requires understanding which part of the environment functionally enables an intended action. 
This capability underlies a wide range of downstream applications, including robotic manipulation~\cite{yamanobe2017brief,nasiriany2025rt}, human--robot interaction~\cite{vallverdu2016emotional,shu2016learning}, and immersive AR/VR systems~\cite{cheng2013affordances,steffen2019framework}. 
Recent advances in 3D scene representations, especially 3D Gaussian Splatting (3DGS)~\cite{kerbl20233d}, have further made scene-level affordance reasoning feasible by providing explicit 3D scene representations with efficient geometry-aware rendering, enabling fine-grained reasoning directly in realistic 3D environments~\cite{wei20253daffordsplat,li2025seqaffordsplat}.

Despite this progress, most existing methods still formulate affordance reasoning as a passive one-shot prediction problem: given a fixed observation, the model directly predicts the target affordance region~\cite{yang2023grounding,li2024laso,zhu2025grounding,wei20253daffordsplat}. 
While effective in visually salient cases, this paradigm often becomes unreliable in complex scenes, where multiple candidates share similar local geometry or the queried affordance lies in a narrow and structurally complex region, as illustrated on the left of Fig.~\ref{fig:teaser}. 
We argue that many such failures are better understood as evidence-limited rather than prediction-limited: the difficulty is not only to infer the target region from the current observation, but also to obtain the task-relevant evidence needed to resolve the ambiguity. 
Following evidence accumulation theories in perception~\cite{gold2007neural}, we treat task-relevant evidence as information that reduces uncertainty about which scene region supports the intended interaction.

A key property of the missing evidence is that it is often cross-dimensional. 
Three-dimensional observations provide precise geometry, spatial relations, and candidate localization, which are essential for preserving scene grounding~\cite{8099499,deng20213d,kerbl20233d}. 
However, geometric evidence alone is often insufficient to distinguish functionally similar regions. 
In contrast, two-dimensional vision foundation models provide richer semantic and functional priors, which are critical for identifying subtle interaction cues but lack direct 3D grounding~\cite{radford2021learning,kirillov2023segment,oquab2023dinov2}. 
Existing methods typically fuse these modalities statically into a task-agnostic representation before prediction~\cite{peng2023openscene,kerr2023lerf,jatavallabhula2023conceptfusion}. 
For fine-grained affordance reasoning, we therefore preserve both modalities and acquire task-relevant evidence on demand, so that geometry and semantics are invoked according to the current ambiguity state rather than fused statically in advance.


Motivated by this observation, and as summarized in Fig.~\ref{fig:teaser}, we reformulate affordance reasoning in 3D Gaussian scenes as a sequential cross-dimensional evidence acquisition problem and propose \textbf{A3R}, an agentic affordance reasoning framework that progressively resolves ambiguity through iterative evidence gathering. 
At each step, the policy is conditioned on a scene-grounded state consisting of the current workspace, the affordance belief, and the action--evidence history.
Conditioned on this state, an MLLM-based policy selects whether to acquire 3D geometric evidence for spatial grounding or 2D semantic evidence for affordance disambiguation.
The acquired evidence is projected back to the scene support and used to update the current affordance belief over the 3DGS scene.
To optimize such sequential decision making, we further introduce a GRPO-based policy learning strategy that improves final reasoning accuracy. 
Extensive experiments on scene-level benchmarks demonstrate that A3R consistently outperforms static one-shot baselines, particularly in unseen affordance cases.





Our main contributions are summarized as follows:
\begin{itemize}
\sloppy
\item We identify insufficient task-relevant evidence under static observation as a key difficulty in fine-grained affordance reasoning, motivating a shift from passive one-shot prediction to sequential evidence acquisition.

\item We propose \textbf{A3R}, an agentic affordance reasoning framework that performs sequential cross-dimensional evidence acquisition by dynamically querying complementary 3D geometric and 2D semantic evidence under an MLLM-based policy.

\item We develop a GRPO-based policy learning strategy that optimizes evidence acquisition trajectories through final task rewards, enabling effective tool selection without step-wise supervision.

\item Extensive experiments demonstrate that A3R consistently outperforms existing static one-shot baselines, achieving an improvement of up to 16.55 mIoU points under the unseen-affordance generalization setting.
\end{itemize}

\section{Related Work}

\subsection{3D Affordance Reasoning}

Affordance reasoning aims to localize the object or region that supports a queried interaction. 
Early work mainly studied this problem in object-centric 3D settings, typically formulating it as dense affordance prediction from static geometric observations~\cite{deng20213d,mo2021where2act}. 
Subsequent studies enriched this formulation with stronger conditioning signals, for example through interaction image cues~\cite{yang2023grounding}, open-vocabulary or language-guided queries~\cite{nguyen2023open,li2024laso,zhu2025grounding}, and MLLMs for affordance reasoning with richer reasoning interfaces~\cite{yu2025seqafford}.

Recent advances in 3DGS have introduced a geometry-aware representation with efficient rendering, making it a practical substrate for fine-grained reasoning in realistic 3D environments~\cite{kerbl20233d}. 
Building on this representation, 3DAffordSplat extends affordance reasoning to 3DGS-based instance-level prediction~\cite{wei20253daffordsplat}, while SeqAffordSplat further studies sequential affordance reasoning in scene-level 3DGS environments~\cite{li2025seqaffordsplat}. 
These methods extend affordance reasoning to richer 3D representations and settings, but they still largely rely on inference from fixed observations.
As a result, they do not explicitly address cases in which a single observation provides insufficient evidence for resolving fine-grained affordance ambiguity in complex 3D scenes.

\subsection{2D--3D Semantic Transfer for Scene Understanding}

Large-scale 2D vision foundation models provide transferable priors for 3D scene understanding, including open-vocabulary semantic features~\cite{radford2021learning}, dense visual features~\cite{oquab2023dinov2}, and segmentation masks~\cite{kirillov2023segment,carion2025sam3}. 
Representative approaches such as OpenScene~\cite{peng2023openscene}, OV3D~\cite{jiang2024open}, and OpenMask3D~\cite{takmaz2023openmask3d} transfer multi-view 2D features or masks into 3D representations through projection and feature lifting, enabling semantic parsing and instance-level understanding in 3D scenes.
Related ideas have also been explored in affordance reasoning, where 2D visual cues or semantic priors are introduced to improve 3D affordance grounding~\cite{yang2023grounding,nguyen2023open,li2025seqaffordsplat}.

This 2D--3D transfer paradigm has further been extended to 3DGS. Recent methods distill 2D features into Gaussian fields, construct semantically enriched 3DGS feature spaces, or support semantic and mask-based understanding directly on 3DGS scenes~\cite{zhou2024feature,qin2024langsplat,cen2025segment,wu2024opengaussian,li2025egosplat}. 
While effective for building semantically enriched 3D representations, these methods typically rely on static 2D--3D fusion, which may propagate view-dependent semantic noise, projection misalignment, and occlusion-induced errors into a fixed 3D feature field~\cite{cen2025tackling,wang2025visibility}.
They are therefore less suited to inference settings that require semantic evidence to be queried under the current ambiguity state.
In contrast, our method treats 2D semantic analysis as an on-demand evidence source for affordance reasoning.

\subsection{Active Perception and Agentic Reasoning}

When fixed observations are insufficient for reliable prediction, an alternative is to acquire additional information during inference~\cite{bajcsy1988active,bajcsy2018revisiting}. This perspective views reasoning as a process of gathering task-relevant evidence under uncertainty. Active perception provides a classical instantiation of this idea in embodied vision and robotics. By selecting informative viewpoints, exploring the environment, or zooming into local regions, these methods improve scene understanding beyond passive recognition from a fixed view~\cite{zhu2025move,liu2026activevla,wang2025affordbot}. 
For example, MTU3D~\cite{zhu2025move} combines visual grounding with exploration in 3D environments, while ActiveVLA~\cite{liu2026activevla} studies active viewpoint control and local refinement for manipulation. 
Among prior works, AffordBot~\cite{wang2025affordbot} is related in that it also moves beyond fixed observation, but does so mainly through instruction-guided selection of an informative rendered view before downstream grounding and motion inference.

Recent advances in LLMs and MLLMs further extend this perspective through agentic reasoning and sequential tool use.
ReAct~\cite{yao2022react} interleaves reasoning and actions to query missing information, Toolformer~\cite{schick2023toolformer} learns tool invocation during inference, and A4-Agent~\cite{zhang2025a4} applies test-time orchestration of pretrained models to zero-shot affordance prediction.

While these developments all move beyond passive inference from fixed observations, existing work typically emphasizes either observation control or general-purpose tool invocation.
Their application to scene-grounded 3D reasoning remains limited, especially in settings where effective reasoning requires coordinated use of complementary 3D geometric and 2D semantic evidence under the current state.

\begin{figure*}
    \centering
    \includegraphics[width=1.0\linewidth]{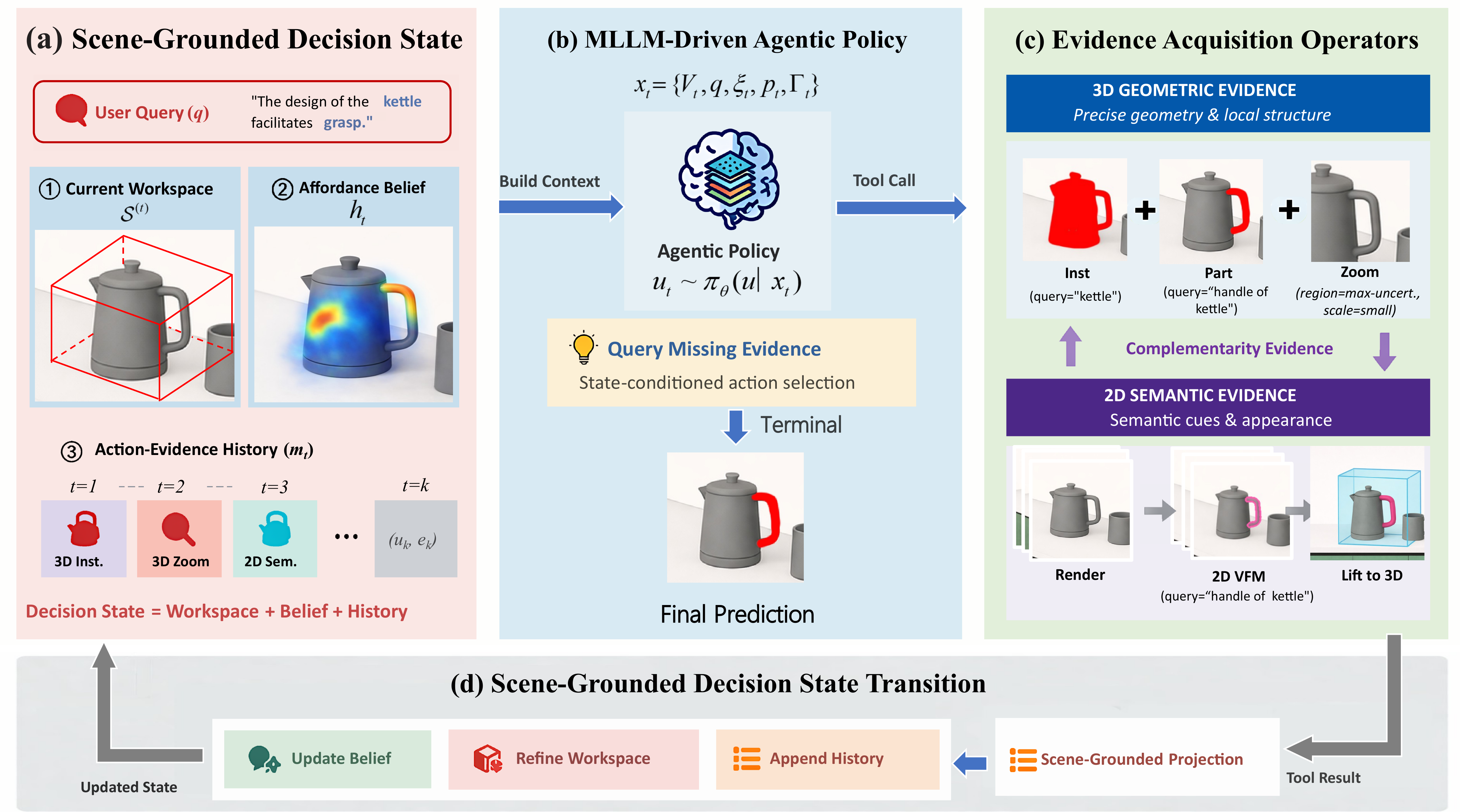}
\caption{
\textbf{Overview of the proposed framework.}
The agent maintains a scene-grounded decision state composed of the current workspace, affordance belief, and action--evidence history.
Conditioned on this state, the policy selects 2D or 3D evidence operators whose outputs are projected back to the scene support to update the state, progressively reducing ambiguity for fine-grained affordance reasoning in 3DGS scenes.
}
    \Description{Overview of the proposed framework for agentic cross-dimensional evidence acquisition in 3DGS scenes.}
    \label{fig:framework_overview}
\end{figure*}



\section{Method}
As illustrated in Fig.~\ref{fig:framework_overview}, we formulate affordance reasoning in 3DGS scenes as a sequential evidence-acquisition process.
Instead of predicting the target region from a fixed observation in one shot, the agent progressively acquires cross-dimensional evidence and updates a scene-grounded affordance belief under the current state.
The overall procedure is summarized in Algorithm~\ref{alg:framework}.

\subsection{Formulation as a Partially Observable Markov Decision Process}

Affordance reasoning in complex 3DGS scenes is inherently partially observable: the queried affordance support is latent, while the agent can only access task-relevant evidence through operator-specific observations acquired during interaction.
We formulate this process as a Partially Observable Markov Decision Process (POMDP)
\[
\bigl(\mathcal{X},\mathcal{U},\mathcal{E},\mathcal{T},\mathcal{O},R\bigr),
\]
where $\mathcal{X}$ is the latent state space, $\mathcal{U}$ the action space, $\mathcal{E}$ the observation space, $\mathcal{T}$ the latent-state transition kernel, $\mathcal{O}$ the observation model, and $R$ the episodic reward.

Given a 3D Gaussian scene $\mathcal{S}=\{g_i\}_{i=1}^{N}$ and an affordance query $q$, we define the latent environment state as a scene-grounded affordance assignment
$
Y=(Y_i)_{i=1}^{N}\in\{0,1\}^{N},
$
where $Y_i=1$ indicates that Gaussian $g_i$ belongs to the queried affordance region. Accordingly, the latent state space is
\begin{equation}
\mathcal{X}=\{0,1\}^{N},
\qquad Y\in\mathcal{X}.
\end{equation}

Since the physical scene and the queried affordance target remain fixed within an episode, the latent state is static across decision steps:
\begin{equation}
\mathcal{T}(Y' \mid Y, u_t)=\mathbf{1}[Y'=Y].
\end{equation}

At step $t$, the agent executes an evidence-acquisition action $u_t\in\mathcal{U}$ on the current workspace $\mathcal{S}^{(t)}$ and receives an observation $e_t\in\mathcal{E}$ drawn from
\begin{equation}
e_t \sim \mathcal{O}(\,\cdot \mid Y, u_t, q, \mathcal{S}^{(t)}).
\end{equation}
Here, $\mathcal{U}$ specifies which operator is executed, while $\mathcal{O}$ specifies the operator-dependent observation generated under the current latent affordance state and workspace.

In principle, optimal inference requires maintaining the posterior belief over the latent state:
\begin{equation}
b_t(Y)=P\!\left(Y \mid e_{1:t},u_{1:t},q,\mathcal{S}\right).
\end{equation}
However, this full posterior is intractable in realistic 3DGS scenes due to the combinatorial state space over $\{0,1\}^{N}$. Instead, we maintain a scene-grounded affordance belief map
\begin{equation}
h_t \in [0,1]^N,
\end{equation}
where $h_t(i)$ denotes the current affordance confidence assigned to Gaussian $g_i$.
Unlike the full posterior $b_t$, $h_t$ is a compact, Gaussian-indexed approximation used for sequential decision-making.

To support policy execution, we further maintain a scene--grounded decision state
\begin{equation}
s_t=\{\mathcal{S}^{(t)},\, h_t,\, m_t\},
\end{equation}
where $\mathcal{S}^{(t)}$ is the current workspace, $h_t$ the current affordance belief map, and $m_t$ the action--evidence history.
Here, $Y$ is the latent environment state, whereas $s_t$ is the agent-maintained internal state for decision-making. In practice, the policy acts on a multimodal context $x_t$ rendered from $s_t$ together with the query.

The objective is to learn a policy that maximizes final affordance grounding quality while avoiding unnecessary evidence-acquisition steps. For a trajectory $\tau=\{u_1,e_1,\dots,u_{T_\tau},e_{T_\tau}\}$ with terminal prediction $\hat{y}$, we optimize
\begin{equation}
\max_{\pi}\;
\mathbb{E}_{(\mathcal{S},q)\sim \mathcal{D}_{\mathrm{env}},\,\tau\sim\pi}
\left[
R(\tau)
\right].
\end{equation}

\subsection{Cross-Dimensional Evidence Acquisition Operators}
\label{sec:operators}

We define a compact action space for querying complementary evidence across dimensions and spatial granularities, covering four main needs in fine-grained affordance reasoning: instance localization, part-level geometric verification, semantic disambiguation, and workspace refinement.

At step $t$, an action is parameterized as
\begin{equation}
u_t=(\tau_t,\rho_t),
\end{equation}
where $\tau_t\in\mathcal{K}$ denotes the operator type and $\rho_t$ its arguments. The action space $\mathcal{U}$ consists of all valid parameterized operator calls together with the terminate action, with
\begin{equation}
\mathcal{K}=\{\texttt{Inst},\texttt{Part},\texttt{SAM2D},\texttt{Zoom},\texttt{Term}\}.
\end{equation}
The observation space is the union of operator-specific evidence spaces,
\begin{equation}
\mathcal{E}=\bigcup_{\tau\in\mathcal{K}}\mathcal{E}_{\tau},
\end{equation}
where $\mathcal{E}_{\tau}$ denotes the structured output space of operator $\tau$. Specifically, the five operators in $\mathcal{K}$ are designed to provide complementary evidence or workspace control under different ambiguity patterns:





\textbf{3D Instance Segmentation (\texttt{Inst})} identifies the query-relevant object instance in the current workspace. It provides object-level geometric evidence that suppresses distractors and narrows subsequent reasoning to the most relevant object support.

\textbf{3D Part Segmentation (\texttt{Part})} predicts affordance-relevant part support in the current 3D workspace. It provides scene-grounded geometric evidence for precise affordance localization.

\textbf{2D Semantic Segmentation with 3D Lifting (\texttt{SAM2D})} extracts semantic evidence from rendered views and lifts it back to the 3D scene. It complements geometric reasoning with semantic cues for affordance disambiguation.

\textbf{3D Zoom (\texttt{Zoom})} restricts the current workspace to a more informative local subscene. It controls the spatial granularity of evidence acquisition for coarse-to-fine reasoning.

\textbf{Terminate (\texttt{Term})} stops evidence acquisition and outputs the current belief as the final prediction.

Additional details are provided in the supplementary material.

\begin{algorithm}[t]
\caption{Sequential Cross-Dimensional Evidence Acquisition for 3DGS Affordance Reasoning}
\label{alg:framework}
\small
\KwIn{3DGS scene $\mathcal{S}$, affordance query $q$, step budget $T$, decision threshold $\delta$}
\KwOut{Affordance prediction $\hat{y}$}

Initialize workspace $\mathcal{S}^{(1)} \leftarrow \mathcal{S}$, belief $h_1$, history $m_1$, and decision state $s_1=\{\mathcal{S}^{(1)},h_1,m_1\}$\;

\For{$t=1$ to $T$}{
    Render visual context $\mathcal{V}_t \leftarrow \mathrm{render}(\mathcal{S}^{(t)}, h_t, m_t)$\;
    Derive textual state summary $\xi_t \leftarrow \mathrm{text}(m_t)$\;
    Construct task prompt $p_t$ and valid operator schema $\Gamma_t$\;
    Form policy context $x_t=\{\mathcal{V}_t, q, \xi_t, p_t, \Gamma_t\}$\;
    Sample action $u_t=(\tau_t,\rho_t)\sim\pi_\theta(\cdot\mid x_t)$\;
    
    \If{$\tau_t=\texttt{Terminate}$}{
        $\hat{y}(i)\leftarrow \mathbf{1}[h_t(i)\ge \delta],\quad i=1,\ldots,N$\;
        \Return{$\hat{y}$}\;
    }

    Execute operator $u_t$ to obtain evidence $e_t$\;
    Project evidence to scene-grounded support: $\bar{e}_t \leftarrow \Pi_{\mathcal{S}}(e_t)$\;
    Update belief: $h_{t+1}\leftarrow g(h_t,\bar{e}_t,\tau_t)$\;
    Update history: $m_{t+1}\leftarrow \Psi(m_t,u_t,e_t)$\;
    Update workspace: $\mathcal{S}^{(t+1)}\leftarrow \Phi(\mathcal{S}^{(t)},u_t,e_t)$\;
    Update decision state: $s_{t+1}\leftarrow \{\mathcal{S}^{(t+1)},h_{t+1},m_{t+1}\}$\;
}

$\hat{y}(i)\leftarrow \mathbf{1}[h_{T+1}(i)\ge \delta],\quad i=1,\ldots,N$\;
\Return{$\hat{y}$}\;
\end{algorithm}

\subsection{MLLM-Driven Agentic Policy}
\label{sec:policy}


To orchestrate sequential evidence acquisition, we model decision-making as a multimodal policy conditioned on the current decision state, and instantiate the policy with an MLLM, since the policy must jointly interpret multimodal context and generate schema-constrained evidence-acquisition actions with flexible, partially open-vocabulary arguments.
Rather than directly predicting the queried affordance region, the policy selects the next evidence-acquisition action from a multimodal context constructed from the current state.

At step $t$, the policy receives a multimodal context consisting of a rendered visual context, the affordance query $q$, the textual component $\xi_t$ of the maintained history, a task prompt $p_t$ specifying operator-use constraints, and a valid operator schema $\Gamma_t$.
The visual input is constructed from multi-view renderings of the current workspace together with visualizations of the current affordance belief and a limited set of historical execution results:
\begin{equation}
\mathcal{V}_t=\mathrm{render}(\mathcal{S}^{(t)}, h_t, m_t).
\end{equation}
The textual input is formed from the text component of the maintained state history:
\begin{equation}
\xi_t=\mathrm{text}(m_t).
\end{equation}
The complete policy context is then given by
\begin{equation}
x_t=\{\mathcal{V}_t,\; q,\; \xi_t,\; p_t,\; \Gamma_t\}.
\end{equation}

Conditioned on $x_t$, the policy predicts the next action
\begin{equation}
u_t \sim \pi_\theta(\cdot \mid x_t),
\end{equation}
where $\pi_\theta$ outputs a schema-constrained operator call $u_t=(\tau_t,\rho_t)$.
Through this formulation, the policy selects the next evidence-acquisition action under the current decision context.

\subsection{Scene-Grounded Decision State Transition}
\label{sec:transition}

After the policy selects an evidence-acquisition action, the returned evidence is incorporated into the scene-grounded decision state through an operator-dependent scene-grounded transition. This transition updates the current belief, history, and workspace according to the role of the selected operator.

Specifically, after executing action $u_t=(\tau_t,\rho_t)$, the returned evidence $e_t$ is projected onto the Gaussian support of the original 3DGS scene:
\begin{equation}
\bar{e}_t=\Pi_{\mathcal{S}}(e_t)\in[0,1]^N,
\end{equation}
where $\Pi_{\mathcal{S}}(\cdot)$ denotes the operator-dependent scene-grounded projection from the returned evidence to the Gaussian-indexed scene support.

The belief update is operator-dependent:
\begin{equation}
h_{t+1}=g(h_t,\bar e_t,\tau_t),
\end{equation}
with
\begin{equation}
g(h_t,\bar e_t,\tau_t)=
\begin{cases}
\bar{e}_t, & \tau_t \in \{\texttt{Part},\texttt{SAM2D}\},\\
h_t, & \tau_t \in \{\texttt{Inst},\texttt{Zoom}\}.
\end{cases}
\label{eq:belief_transition}
\end{equation}
That is, affordance-predictive operators update the current belief through their projected scene-grounded outputs, whereas workspace-control operators refine the current workspace while leaving the belief unchanged. This direct replacement design avoids committing the decision state to noisy intermediate masks and allows later evidence to revise the current affordance belief cleanly.

The history and workspace are updated as
\begin{equation}
m_{t+1}=\Psi(m_t,u_t,e_t),
\qquad
\mathcal{S}^{(t+1)}=\Phi(\mathcal{S}^{(t)},u_t,e_t),
\end{equation}
where implementation details of $\Psi(\cdot)$ and $\Phi(\cdot)$ are deferred to the supplementary material.

When the policy selects $\texttt{Term}$, the final prediction is obtained by thresholding the current belief:
\begin{equation}
\hat{y}(i)=\mathbf{1}[h_t(i)\ge\delta], \quad i=1,\ldots,N,
\end{equation}
where $\delta$ is the decision threshold. 
If termination is not triggered before the maximum interaction budget $T$ is reached, the episode stops after the final state update, and we output the prediction obtained by thresholding the last maintained belief $h_{T+1}$.

\subsection{GRPO-Based Policy Learning}
\label{sec:grpo}

We optimize the policy $\pi_\theta$ by maximizing the expected episodic reward over sequential evidence-acquisition trajectories. Since the utility of an evidence-acquisition action is often only reflected in the final affordance prediction after multiple interaction steps, we adopt Group Relative Policy Optimization (GRPO)~\cite{shao2024deepseekmath} to train the MLLM policy from on-policy rollouts over complete trajectories.

Given a sampled scene $\mathcal{S}$ and query $q$, the current policy rolls out a group of $G$ evidence-acquisition trajectories $\mathcal{G}=\{\tau_1,\dots,\tau_G\}$ up to length $T$. Each trajectory $\tau_i$ consists of a sequence of policy contexts, selected actions, and returned evidence over $T_i$ interaction steps, produces a terminal prediction $\hat{y}_i$, and receives the reward
\begin{equation}
R(\tau_i)=R_{\mathrm{acc}}(\hat{y}_i,y^\ast)+R_{\mathrm{step}}(T_i),
\end{equation}
where the accuracy reward is defined as
\begin{equation}
\begin{aligned}
R_{\mathrm{acc}}(\hat{y}_i,y^\ast)
&=\lambda_{\mathrm{iou}}\,\mathrm{IoU}(\hat{y}_i,y^\ast)
+\lambda_{\mathrm{succ}}\,\mathbf{1}[\mathrm{IoU}(\hat{y}_i,y^\ast)\ge\eta] \\
&\quad-\lambda_{\mathrm{fail}}\,\mathbf{1}[\mathrm{IoU}(\hat{y}_i,y^\ast)<\epsilon],
\end{aligned}
\end{equation}
and the step penalty is
\begin{equation}
R_{\mathrm{step}}(T_i)=-\lambda_{\mathrm{step}}\,T_i.
\end{equation}

GRPO computes a group-relative advantage by normalizing rewards within the rollout group:
\begin{equation}
A_i=\frac{R(\tau_i)-\mu_{\mathcal{G}}}{\sigma_{\mathcal{G}}+\varepsilon},
\end{equation}
where $\mu_{\mathcal{G}}$ and $\sigma_{\mathcal{G}}$ denote the empirical mean and standard deviation of rewards in $\mathcal{G}$.

The policy is optimized with the clipped GRPO objective regularized by a KL penalty to a reference policy $\pi_{\mathrm{ref}}$. For each step in trajectory $\tau_i$, we define the importance ratio
\begin{equation}
\rho_{i,t}(\theta)=
\frac{\pi_\theta(u_{i,t}\mid x_{i,t})}
{\pi_{\theta_{\mathrm{old}}}(u_{i,t}\mid x_{i,t})},
\end{equation}
and optimize
\begin{equation}
J_{\mathrm{GRPO}}(\theta)
=
\mathbb{E}_{q,\mathcal{S}\sim\mathcal{D}_{\mathrm{env}}}
\left[
\frac{1}{G}\sum_{i=1}^{G}\sum_{t=1}^{T_i}
L_{i,t}(\theta)
-\beta\,\mathbb{D}_{\mathrm{KL}}(\pi_\theta \parallel \pi_{\mathrm{ref}})
\right],
\end{equation}
where
\begin{equation}
L_{i,t}(\theta)=
\min\big(
\rho_{i,t}(\theta)A_i,\,
\mathrm{clip}(\rho_{i,t}(\theta),1-\kappa,1+\kappa)A_i
\big).
\end{equation}

All actions within the same trajectory share the same trajectory-level advantage $A_i$, encouraging operator choices that lead to better final outcomes. In practice, we optimize the MLLM policy with LoRA while keeping the underlying evidence operators frozen.

\section{Experiments}
\subsection{Experimental Setup}

\noindent\textbf{Datasets and protocol.}
We evaluate our method on two 3DGS-based affordance benchmarks, 3DAffordSplat~\cite{wei20253daffordsplat} and SeqAffordSplat~\cite{li2025seqaffordsplat}. 
3DAffordSplat is instance-level, while SeqAffordSplat is scene-level. 
For SeqAffordSplat, we use the Single setting and further partition it into Seen and UnSeen splits following 3DAffordSplat for a consistent generalization protocol.

\noindent\textbf{Evaluation metrics.}
We report standard affordance grounding metrics, including mIoU, AUC, and SIM, together with the average number of interaction steps for reasoning efficiency.

\noindent\textbf{Compared methods.}
We compare against representative static affordance reasoning baselines, including IAGNet~\cite{yang2023grounding}, PointRefer~\cite{li2024laso}, and AffordSplatNet~\cite{wei20253daffordsplat}. 
We also compare with internal variants, including fixed acquisition strategies and different policy backbones. 
Unless otherwise stated, studies on evidence components and policy backbones use the training-free policy.

\noindent\textbf{Implementation and training details.}
The policy selects from four evidence-acquisition operators: \texttt{Part}, \texttt{Inst}, \texttt{Zoom}, and \texttt{SAM2D}. The \texttt{Part} and \texttt{Inst} operators are implemented based on AffordSplatNet~\cite{wei20253daffordsplat}. The \texttt{SAM2D} operator renders 6 upper-hemisphere views, applies SAM3-based~\cite{carion2025sam3} segmentation, and lifts the results back to the 3D Gaussian support. In the main experiments, we use Qwen3-VL-8B as the policy model and optimize it with GRPO while keeping the underlying evidence operators frozen. LoRA is applied to the attention projections. More details are deferred to the supplementary material.

\subsection{Main Results}

Tab.~\ref{tab:main_seqaffordsplat} and Tab.~\ref{tab:main_3daffordsplat} report the main results. 
Our method consistently outperforms all static baselines on both datasets under both Seen and Unseen settings, with especially large gains on the more challenging Unseen splits.

In these cases, static predictors must decide from incomplete evidence, while our method can progressively acquire the missing evidence needed to disambiguate functionally similar candidates. 
The improvement is more pronounced on SeqAffordSplat, where scene-level ambiguity is stronger and a single observation is often insufficient. In these cases, static predictors must decide from incomplete evidence, while our method can progressively acquire the missing evidence needed to disambiguate functionally similar candidates. 
This advantage becomes more pronounced on the Unseen splits, where reliable prediction depends less on familiar training patterns and more on resolving ambiguity through additional evidence acquisition.

Notably, the training-free version already surpasses prior one-shot baselines by a clear margin, showing that the proposed formulation is effective even without policy learning. GRPO further strengthens the framework by learning more effective evidence-acquisition decisions under a limited step budget.

\begin{table}[htbp]
\centering
\small
\caption{Main results on SeqAffordSplat.}
\begin{tabular}{llccc}
\toprule
Setting & Method & mIoU$\uparrow$ & AUC$\uparrow$ & SIM$\uparrow$ \\
\midrule
\multirow{5}{*}{Seen}
& IAGNet~\cite{yang2023grounding}          & 9.70  & 79.45 & 0.168 \\
& PointRefer~\cite{li2024laso}             & 17.61 & 85.90 & 0.260 \\
& AffordSplatNet~\cite{wei20253daffordsplat}& 21.04 & 83.37 & 0.283 \\
& Ours (training-free)                     & 26.81 & 86.11 & 0.330 \\
& Ours (RL)                                & \textbf{28.28} & \textbf{88.51} & \textbf{0.355} \\
\midrule
\multirow{5}{*}{Unseen}
& IAGNet~\cite{yang2023grounding}          & 10.48 & 61.05 & 0.176 \\
& PointRefer~\cite{li2024laso}             & 9.28  & 72.27 & 0.174 \\
& AffordSplatNet~\cite{wei20253daffordsplat}& 9.01  & 71.34 & 0.155 \\
& Ours (training-free)                     & 22.62 & 85.48 & 0.259 \\
& Ours (RL)                                & \textbf{25.56} & \textbf{86.33} & \textbf{0.314} \\
\bottomrule
\end{tabular}
\label{tab:main_seqaffordsplat}
\end{table}

\begin{table}[htbp]
\centering
\small
\caption{Main results on 3DAffordSplat.}
\label{tab:main_3daffordsplat}
\begin{tabular}{llccc}
\toprule
Setting & Method & mIoU$\uparrow$ & AUC$\uparrow$ & SIM$\uparrow$ \\
\midrule
\multirow{5}{*}{Seen}
& IAGNet~\cite{yang2023grounding}          & 14.63 & 56.67 & 0.350 \\
& PointRefer~\cite{li2024laso}             & 18.40 & 78.50 & 0.430 \\
& AffordSplatNet~\cite{wei20253daffordsplat}& 30.25 & 83.85 & 0.440 \\
& Ours (training-free)                     & 34.88 & 81.86 & 0.462 \\
& Ours (RL)                                & \textbf{36.30} & \textbf{83.90} & \textbf{0.475} \\
\midrule
\multirow{5}{*}{Unseen}
& IAGNet~\cite{yang2023grounding}          & 4.70  & 40.77 & 0.240 \\
& PointRefer~\cite{li2024laso}             & 15.90 & 67.00 & 0.310 \\
& AffordSplatNet~\cite{wei20253daffordsplat}& 17.31 & 67.18 & 0.320 \\
& Ours (training-free)                     & 24.22 & 70.25 & 0.322 \\
& Ours (RL)                                & \textbf{26.27} & \textbf{71.95} & \textbf{0.328} \\
\bottomrule
\end{tabular}
\end{table}

\subsection{Comparison of Evidence Acquisition Strategies}

Tab.~\ref{tab:decision_strategy} compares different evidence acquisition strategies under the same operator set and backbone, while evaluating sequential methods under a shared interaction budget. 
Using only 3D geometric evidence performs better than using only 2D semantic evidence, indicating that scene-grounded geometry provides the primary basis for affordance localization in 3DGS scenes. 
However, both single-source variants remain clearly inferior to the full model, showing that neither geometry nor semantics alone is sufficient in ambiguity-heavy cases.

Static 2D--3D fusion also underperforms our framework, suggesting that the key limitation is not the absence of cross-dimensional information itself, but the inability of static fusion to acquire task-relevant evidence conditioned on the current ambiguity state. 

Compared with hand-designed fixed-sequence baselines and the uncertainty-driven heuristic, our policy achieves the best overall performance by adaptively selecting both the evidence source and the execution order under the current scene-grounded state.

\begin{figure}[t]
    \centering
    \includegraphics[width=\linewidth]{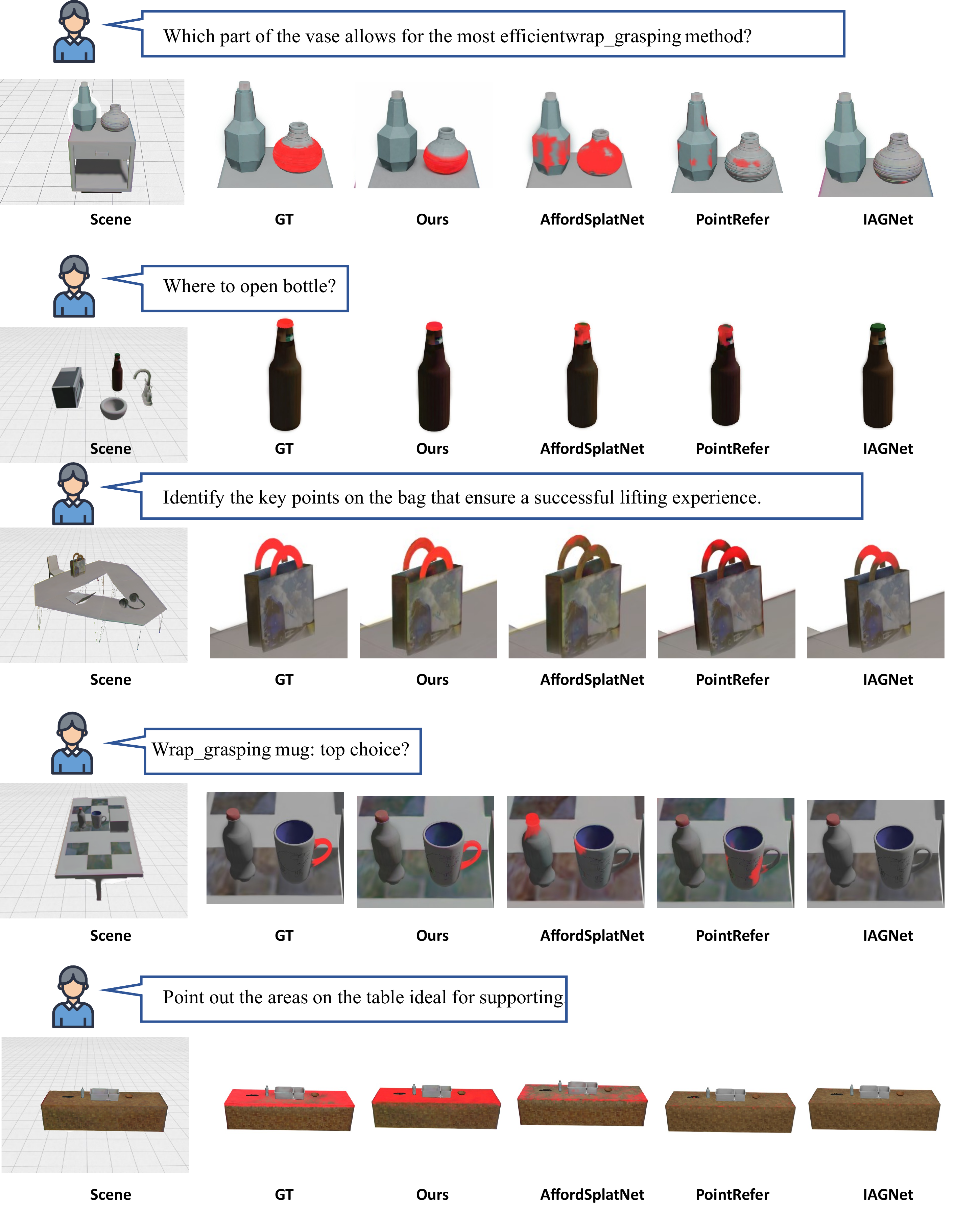}
    \caption{Qualitative comparison with static baselines. Our method produces more accurate and spatially precise affordance predictions, especially in ambiguity-heavy scenes that require fine-grained localization and distractor suppression.}
    \label{fig:results}
    \Description{Qualitative comparison with static baselines for affordance reasoning.}
\end{figure}

\begin{figure*}[t]
    \centering
    \includegraphics[width=0.9\linewidth]{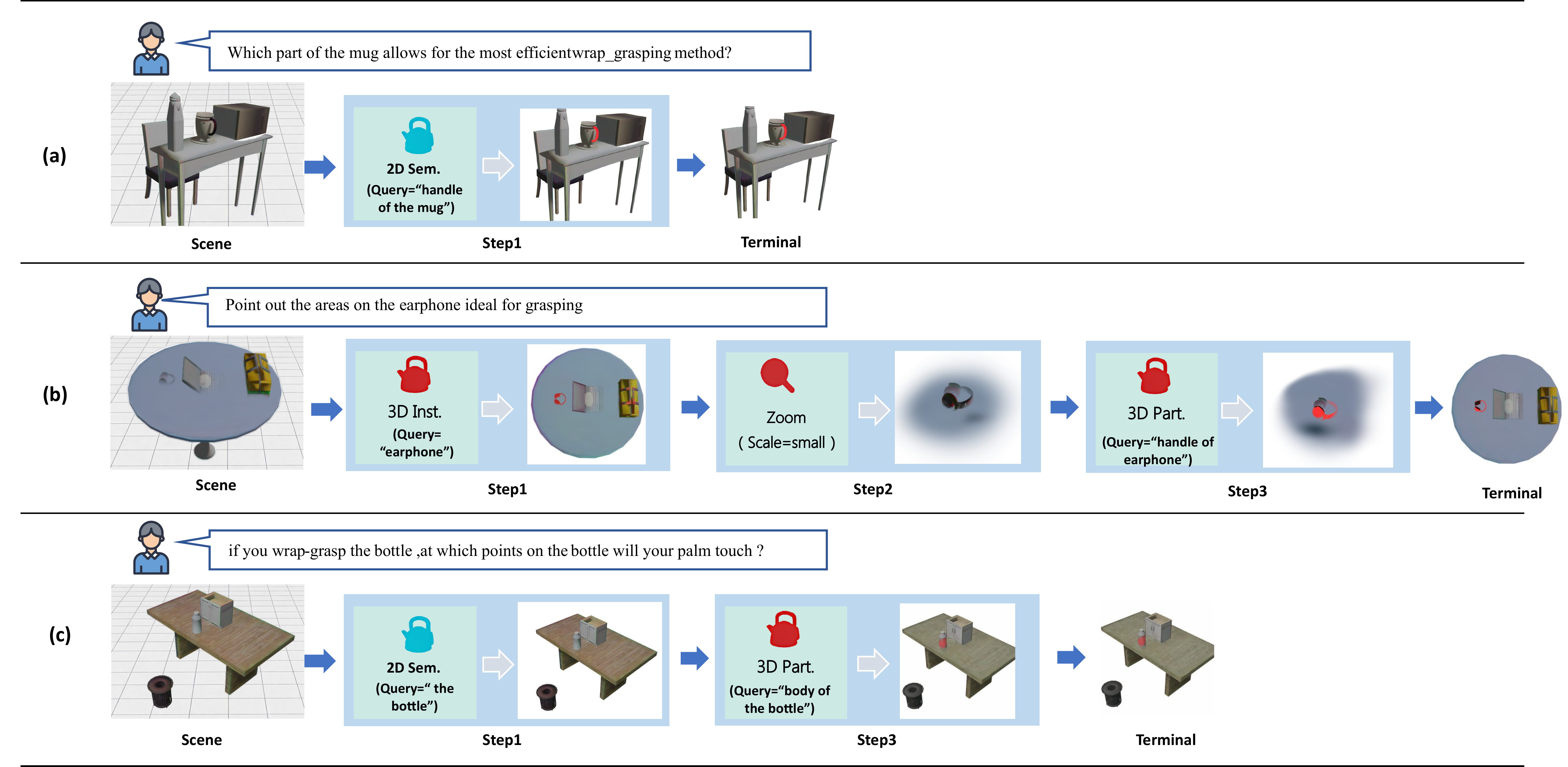}
    \caption{Visualization of evidence-acquisition trajectories. The agent adaptively selects 2D semantic and 3D geometric operators based on the current belief, progressively refining the affordance prediction through coarse-to-fine cross-dimensional evidence acquisition.}
    \label{fig:evidence-acquisition}
    \Description{Visualization of evidence-acquisition trajectories for agentic cross-dimensional reasoning.}
\end{figure*}

\begin{table}[htbp]
\centering
\small
\caption{Comparison of evidence acquisition strategies. All methods use the same evidence operators and the same backbone.}
\label{tab:decision_strategy}
\begin{tabular}{lcccc}
\toprule
Strategy & mIoU$\uparrow$ & AUC$\uparrow$ & SIM$\uparrow$ & Avg. Steps$\downarrow$ \\
\midrule
SAM2D only & 18.40 & 72.20 & 0.230 & 1 \\
Part only & 21.04 & 83.37 & 0.283 & 1 \\
\midrule
Static fusion & 19.10 & 82.89 & 0.267 & 1 \\
\midrule
Fixed Seq. A & 21.97 & 82.01 & 0.302 & 3 \\
Fixed Seq. B & 18.96 & 85.39 & 0.212 & 2 \\
Fixed Seq. C & 21.80 & 84.21 & 0.248 & 3 \\
Fixed Seq. D & 18.13 & 84.59 & 0.203 & 3 \\

\midrule
Greedy policy & 19.17 & 84.92 & 0.214 & 3.19 \\
\midrule
Ours (training-free) & 26.81 & 86.11 & 0.330 & 2.66 \\
Ours (RL) & \textbf{28.28} & \textbf{88.51} & \textbf{0.355} & \textbf{2.60} \\
\bottomrule
\end{tabular}

\vspace{2pt}
\begin{minipage}{0.98\linewidth}
\footnotesize
\textbf{Notes on baseline strategies:} 
\textbf{Static fusion} augments 3D Gaussian attributes with PCA-reduced 2D SAM features for one-shot prediction. 
\textbf{Fixed sequences} instantiate natural hand-designed workflows prior to termination:
(A) \textit{Coarse-to-fine geometric refinement}: 3D part segmentation $\rightarrow$ 3D zoom $\rightarrow$ 3D part segmentation; 
(B) \textit{Instance-first localization}: 3D instance segmentation $\rightarrow$ 3D part segmentation; 
(C) \textit{Geometry-semantics mixed reasoning}: 3D instance segmentation $\rightarrow$ 3D zoom $\rightarrow$ 2D SAM segmentation; 
(D) \textit{Instance-first coarse-to-fine geometric reasoning}: 3D instance segmentation $\rightarrow$ 3D zoom $\rightarrow$ 3D part segmentation.
\textbf{Greedy policy} adaptively selects the next operator based on the uncertainty of the current affordance belief. 
\end{minipage}

\end{table}

\begin{table}[htbp]
\centering
\small
\caption{Ablation study of the proposed evidence components.}
\label{tab:component_ablation}
\begin{tabular}{ccccc|ccc}
\toprule
No. & SAM2D & Part & Inst & Zoom & mIoU$\uparrow$ & AUC$\uparrow$ & SIM$\uparrow$ \\
\midrule
1 &  & \checkmark &  &  & 21.04 & 83.37 & 0.28 \\
2 & \checkmark &  &  &  & 18.40 & 72.20 & 0.23 \\
3 & \checkmark & \checkmark &  &  & 23.04 & 84.55 & 0.29 \\
4 & \checkmark & \checkmark & \checkmark &  & 26.11 & 86.01 & 0.31 \\
5 & \checkmark & \checkmark & \checkmark & \checkmark & \textbf{26.81} & \textbf{86.11} & \textbf{0.33} \\
\bottomrule
\end{tabular}
\end{table}

\subsection{Ablation on Evidence Components}

Tab.~\ref{tab:component_ablation} evaluates the contribution of each evidence component under the same interaction budget. Using only Part outperforms using only SAM2D, indicating that accurate affordance localization in 3DGS scenes primarily depends on scene-grounded geometric evidence. Combining SAM2D with Part further improves all metrics, showing that 2D semantic and 3D geometric evidence are complementary in ambiguity-heavy cases.

Adding Inst brings a further gain by suppressing irrelevant scene content before finer reasoning, making subsequent evidence acquisition more focused. Adding Zoom yields the best overall performance, indicating that controlling the spatial granularity of evidence acquisition is also important for fine-grained affordance localization.

Overall, the gain comes from coordinating complementary evidence sources with workspace refinement, rather than relying on any single component alone.



\subsection{Effect of Policy Backbone}

Tab.~\ref{tab:qwen_backbone} compares different MLLM backbones under the same interaction budget. Larger backbones consistently improve both grounding accuracy and reasoning efficiency. Qwen3-VL-8B achieves the best overall performance with the fewest average steps, indicating that stronger policy models make more effective evidence-acquisition decisions from the current multimodal context, rather than relying on more interactions.

\begin{table}[htbp]
\centering
\small
\caption{Effect of agent backbone choice.}
\label{tab:qwen_backbone}
\begin{tabular}{lcccc}
\toprule
Agent Backbone & mIoU$\uparrow$ & AUC$\uparrow$ & SIM$\uparrow$ & Avg. Steps$\downarrow$ \\
\midrule
Qwen3-VL-2B  & 19.98 & 75.56 & 0.24 & 4.96 \\
Qwen3-VL-4B  & 22.63 & 85.00 & 0.28 & 3.92 \\
Qwen3-VL-8B  & \textbf{26.81} & \textbf{86.11} & \textbf{0.33} & \textbf{2.66} \\
\bottomrule
\end{tabular}
\end{table}

\subsection{Qualitative Results}
\noindent\textbf{Comparison with static baselines.}
As shown in Fig.~\ref{fig:results}, static methods often fail in ambiguity-heavy scenarios, either activating multiple candidates or producing coarse predictions. 
This is particularly evident when geometrically similar regions coexist or when fine-grained affordance boundaries are required. 
In contrast, our method consistently focuses on the correct region with sharper and more localized predictions. 
This behavior reflects its ability to acquire missing task-relevant evidence, rather than relying on a single fixed observation.


\noindent\textbf{Visualization of evidence-acquisition trajectories.}
Fig.~\ref{fig:evidence-acquisition} shows how the agent progressively resolves ambiguity through cross-dimensional evidence acquisition. The selected operator sequence varies with the current decision state: simpler cases can be resolved with an early 2D semantic query, whereas harder cases require multi-step reasoning with instance grounding, workspace zoom, and part-level verification. A clear coarse-to-fine pattern emerges across the trajectories, where 3D operators improve spatial grounding and 2D operators provide complementary semantic cues. These examples illustrate that the performance gain comes from adaptive evidence acquisition and progressive refinement under the current state, rather than from stronger one-shot prediction alone.


\section{Conclusion}
In this paper, we revisited fine-grained affordance reasoning in 3DGS scenes and argued that many challenging cases are fundamentally evidence-limited under static observation. 
Based on this insight, we proposed \textbf{A3R}, which reformulates the task as a sequential cross-dimensional evidence acquisition problem. 
Extensive experiments on 3DGS-based affordance benchmarks demonstrate that A3R consistently outperforms static one-shot baselines, with especially clear gains under unseen-affordance generalization. 
Overall, our results suggest that fine-grained affordance reasoning in complex 3D scenes benefits from moving beyond fixed-observation prediction toward adaptive evidence acquisition.


\bibliographystyle{ACM-Reference-Format}
\bibliography{ref}

\end{document}